\pdfoutput=1

\documentclass[11pt]{article}
\usepackage{multicol}
\usepackage{wrapfig}
\usepackage{hyperref}
\usepackage{caption}
\usepackage[]{acl2023}

\usepackage{array,booktabs,ragged2e}
\usepackage{multirow}
\usepackage{times}
\usepackage{latexsym}
\usepackage{arydshln}
\usepackage{pifont}
\newcommand{\checkmark}{\ding{51}}%
\newcommand{\xmark}{\ding{55}}%
\usepackage{float}

\usepackage{placeins}
\usepackage{afterpage}
\usepackage{amsfonts}
\usepackage[T1]{fontenc}

\usepackage[utf8]{inputenc}

\newcommand{\drop}[1]{{\color[HTML]{CB4335}(-#1)}}
\newcommand{\up}[1]{{\color[HTML]{2E86C1}(+#1)}}
\newcommand{\improve}[1]{{\color[HTML]{2E86C1}+}}

\usepackage{microtype}
\usepackage{graphicx}

\usepackage{amsmath}
\DeclareMathOperator{\adv}{Adv}

\usepackage{inconsolata}

\renewcommand\vec{\mathbf}

\newcommand{\treelogo}{\raisebox{5pt}{\includegraphics[scale=0.050]{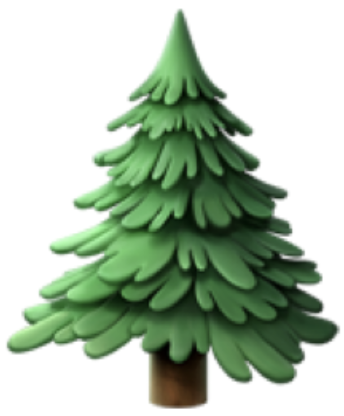}}}
\newcommand{\gtlogo}{\raisebox{3.4pt}{\includegraphics[scale=0.025]{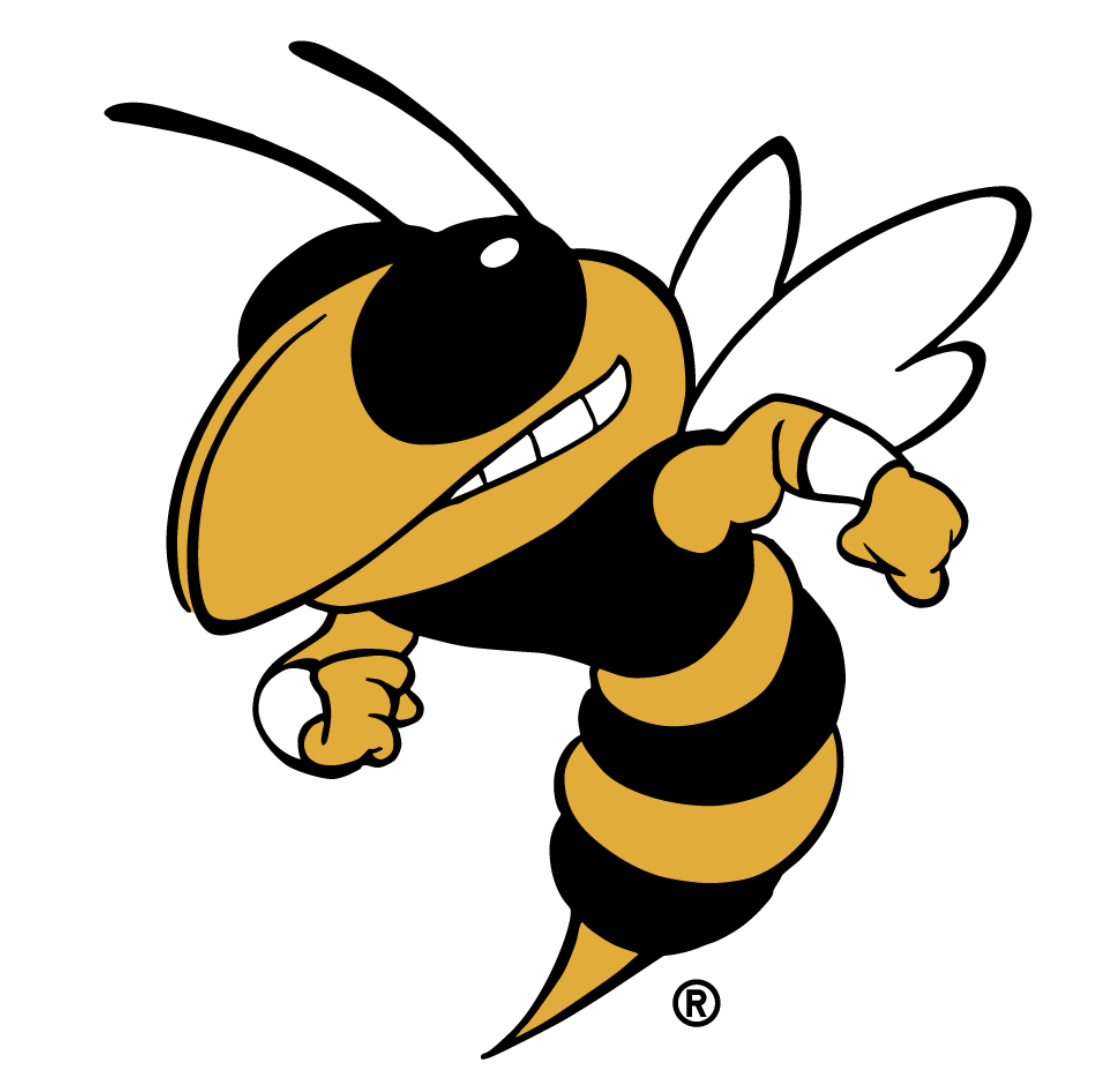}}}

\title{TADA: Task-Agnostic Dialect Adapters for English}

\author{William Held\gtlogo\hspace{10pt} Caleb Ziems\treelogo\hspace{10pt}Diyi Yang\treelogo
\\
  \gtlogo Georgia Institute of Technology, \treelogo Stanford University\\
  \texttt{wheld3@gatech.edu}
  }
\begin{document}
\maketitle
\begin{abstract}
Large Language Models, the dominant starting point for Natural Language Processing (NLP) applications, fail at a higher rate for speakers of English dialects other than Standard American English (SAE). Prior work addresses this using task-specific data or synthetic data augmentation, both of which require intervention for each dialect and task pair. This poses a scalability issue that prevents the broad adoption of robust dialectal English NLP. We introduce a simple yet effective method for task-agnostic dialect adaptation by aligning non-SAE dialects using adapters and composing them with task-specific adapters from SAE. \textbf{T}ask-\textbf{A}gnostic \textbf{D}ialect \textbf{A}dapters (\textbf{TADA}) improve dialectal robustness on 4 dialectal variants of the GLUE benchmark without task-specific supervision.\footnote{We release code for training both traditional and task-agnostic adapters for English dialects on \href{https://github.com/Helw150/tada}{GitHub}
and finetuned models, adapters, and TADA modules on \href{https://huggingface.co/models?sort=downloads\&search=WillHeld+tada}{HuggingFace}.}
\end{abstract}

\section{Introduction}

Large Pretrained Language Models~\citep[LLMs;][]{bert, roberta, t5} have been shown to perform much worse for English dialects other than Standard American English (SAE)~\citep{value, multivalue}. Existing work on dialectal English NLP is task-specific, using manually annotated dialect data~\citep{blodgett-etal-2018-twitter, gang-related}, weak-supervision~\citep{pos-tagger, socially-equitable}, or data augmentation~\citep{value, multivalue}. 

As LLMs become a general-purpose technology, they are applied in an increasing number of scenarios by users who are not formally trained in Machine Learning~\citep{opps-and-risk}. Non-experts rarely look beyond accuracy~\citep{non-experts}, making them less likely to value robustness above the cost of training~\citep{ethayarajh-jurafsky-2020-utility}. Unmitigated dialect bias in this long tail of tasks has the potential to exacerbate harms due to unfair allocation of resources~\citep{bender2021dangers}. 

Dialectal discrepancies originate in biases in the filtering of LLM pretraining data before finetuning~\citep{whose-language}. Despite dialects being definitionally similar, training which enables task-agnostic zero-shot transfer is underexplored relative to potential utility~\citep{local-languages}. Such task-agnostic transfer methods are natural, practical, and offer a scalable solution for English dialects across the growing spectrum of NLP applications.

\begin{figure}[t]
\centering
    \includegraphics[width=0.99\columnwidth]{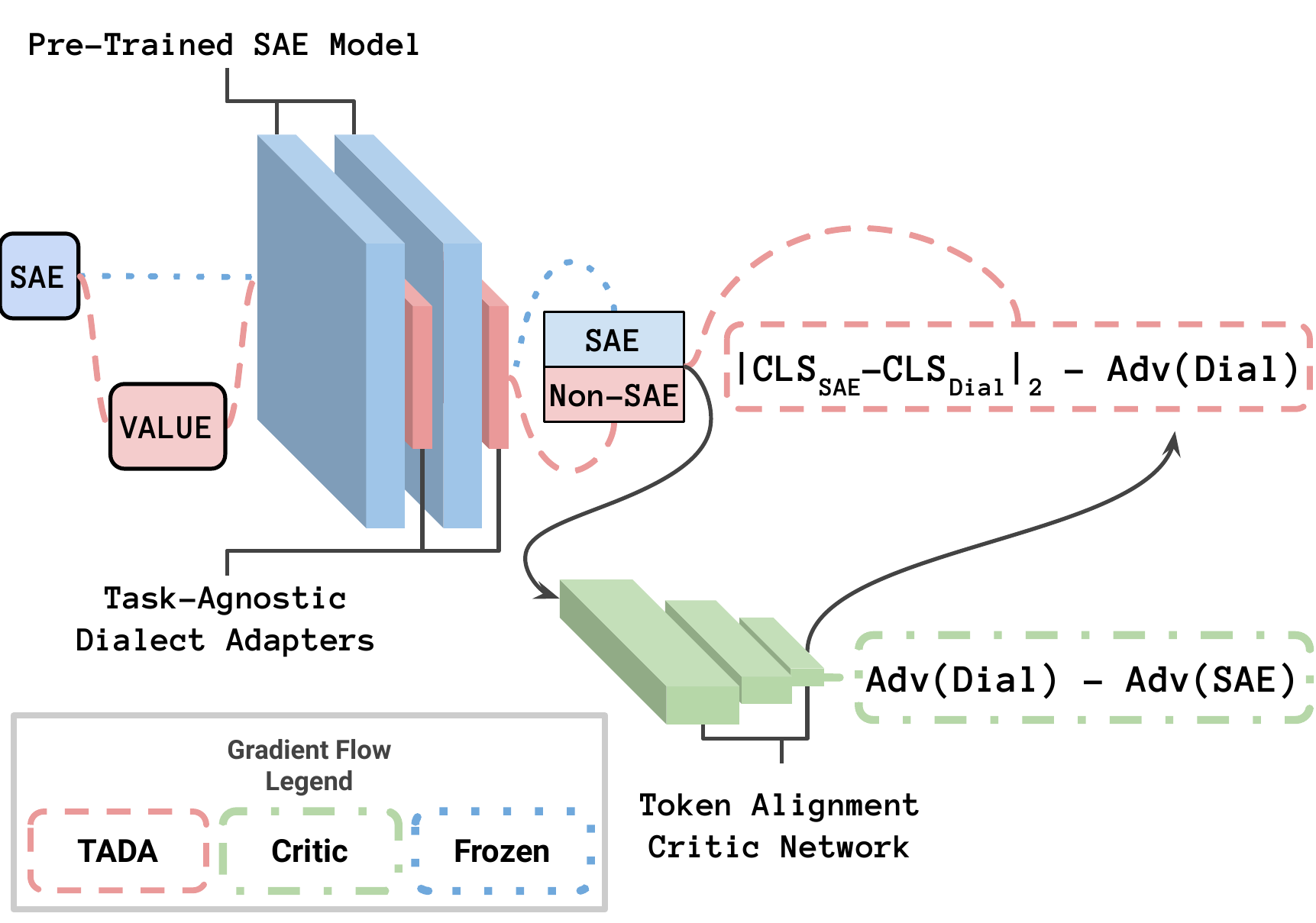}
    \caption{TADA trains adapters with both sequence and token level alignment loss between SAE and a target dialect. When stacked before task-specific SAE adapters, TADA provides dialect robustness for the target task.}
    \label{fig:pipeline}
\end{figure}

This work contributes the first pursuit of these goals with \textbf{T}ask-\textbf{A}gnostic \textbf{D}ialect \textbf{A}dapters (\textbf{TADA}). Adapters, bottlenecks placed between transformer layers, provide a parameter-efficient~\citep{houlsby2019parameter} and composable~\citep{mad-x} foundation for task-agnostic dialect adaptation, given the low-resourced nature of most dialects. As shown in Figure \ref{fig:pipeline}, TADA modules are trained to align non-SAE dialect inputs with SAE inputs at multiple levels with both a sequence-level contrastive loss and a novel morphosyntactic loss. 

We show the empirical effectiveness of TADA on 4 dialect variants of GLUE~\citep{Wang2018GLUEAM} with perturbations from~\citet{multivalue}. We release TADA as a plug-and-play tool for mitigating dialect discrepancies, launching a scalable pathway to dialect-inclusive English NLP. 

\section{Related Work}
\paragraph{NLP For English Dialects} 
Existing work on NLP for English dialects has largely focused on data collection and weak supervision. \citet{pos-tagger} uses online lexicons to provide weak supervision for AAE. \citet{gang-related} manually annotates a small dataset and uses domain adaptation methods to enable transfer. \citet{socially-equitable} collects a geographically diverse set of English data and uses distant supervision signals to annotate a large and representative language ID corpus. Multi-VALUE~\citep{value, multivalue} develops a data augmentation framework for task-specific training in many common English dialects. Our work proposes a complementary task-agnostic intervention for English NLP. 

\paragraph{Cross-Lingual Alignment}
Cross-lingual alignment has become a common approach for task-agnostic zero-shot transfer across languages. Explicit lexical alignment can be used to learn cross-lingual word embeddings for downstream tasks~\citep{cross-lingual-foundation, crossling-embed, robust-alignment, wasserstein-procrustes}. More recent work shows that end-to-end models can implicitly learn to align representations~\citep{low-resource-translation, translation-lm, xlm-r, mt5}. These alignment methods often perform better on highly similar languages, making them theoretically well-suited for dialects. By using explicit alignment with composable modules, our work is the first to explore such techniques for English dialectal NLP.

\paragraph{Adapters}
A growing body of research has been devoted to finding scalable methods for adapting increasingly large-scale pre-trained models. \citet{houlsby2019parameter} adapt large models using bottleneck layers (with skip-connection) between each layer. This idea has been extended in many domains~\citep{stickland2019bert,pfeiffer-etal-2021-adapterfusion,rebuffi2017learning,lin-etal-2020-exploring}. Most relevant, \citet{mad-x} showed that discrete language modeling adapters and task adapters can be composed for effective cross-lingual multi-task transfer. Our experiments exploit specialized dialectal data augmentation to extend this approach to English dialects using explicit alignment loss.

\begin{table*}
\resizebox{0.99\textwidth}{!}{%
\begin{tabular}{cccc|c|c|c|c|c|c|c|c}
             \multicolumn{4}{c|}{Dialect Adaptation Details} &           \multicolumn{8}{c}{AAE Glue Performance}               \\ \hline
Approach & Method    & Task-Agnostic  & Dialect Params. & COLA & MNLI & QNLI & RTE  & QQP  & SST2 & STS-B & Mean      \\
\hline
N/A                & Finetuning & \checkmark & 0                   & 13.5 & 82.0 & 89.3 & 71.8 & 87.1 & 92.0 & 89.9  & 75.1                 \\
N/A                & Adapters   & \checkmark & 0                      & 14.1 & 83.7 & 90.3 & 67.1 & 86.8 & 92.1 & 88.7  & 74.7                 \\ \hdashline
VALUE       & Finetuning & \xmark &$T\times 110M$                 & 19.8 & 84.9 & 90.8 & 74.4 & 89.6 & 92.4 & 90.9  & 77.5                 \\
VALUE       & Adapters   & \xmark & $T\times 895K$                    & 40.2 & 85.8 & 92.2 & 73.6 &   89.7   & 93.6 & 90.3  & 80.8 \\
\hdashline
TADA               & Adapters  & \checkmark & $895K$                  & 29.5\improve{} & 84.8\improve{} & 91.7\improve{} & 67.2\improve{} & 88.1\improve{} & 91.9 & 89.6\improve{}  & 77.5\improve{}                
\end{tabular}
}
\caption{\textbf{Dialect Adaptation GLUE} results of RoBERTa Base~\citep{roberta} for the 7 GLUE Tasks (Matthew's Corr. for CoLA; Pearson-Spearman Corr. for STS-B; Accuracy for all others). $T$  is the number of target tasks for dialect adaptation.  Tasks where TADA improves the performance of task-specific SAE adapters, are marked with \improve{}.}
\label{main-table}
\end{table*}

\section{TADA: Task-Agnostic Dialect Adapters}
As an initial effort, 
TADA aims to provide task-agnostic dialect robustness for English NLP. To do so, we build on work from both multilingual NLP and computer vision and apply explicit alignment losses for transfer learning. 
Concretely, we first generate a synthetic sentence-parallel corpus using the morphosyntactic transformations created by \citet{multivalue}. Using these parallel sentences, we train TADA to align using a contrastive loss at the sequence level and an adversarial loss at the token level. At test time, TADA modules are stacked with task-specific adapters trained on SAE to improve the dialect performance on the target task without further training.

\subsection{Synthetic Parallel Data}\label{synth}
While cross-lingual transfer has leveraged the wealth of sentence parallel bi-texts from machine translation to learn alignment, there are no large-scale parallel English dialectal datasets. Therefore, we leverage Multi-VALUE, a rule-based morphosyntactic SAE to a non-SAE translation system to create parallel data~\citep{multivalue}. 

We start with SAE sentences sampled from the Word-in-Context (WiC) Dataset~\citep{WiC}. WiC is designed to contain lexically diverse sentences and is sourced from high-quality lexicographer written examples~\citep{wordnet, verbnet}. This avoids our alignment modules overfitting to specific vocabulary or noise from low-quality examples. We generate 1,000 such pairs, an amount which could be feasibly replaced with human-translated data. 

This data limitation is intentional, as Multi-VALUE could alternatively used to do large-scale pretraining on transformed data~\citep{qian2022perturbation}. With smaller data limitations, the data used to train TADA can be manually curated native speakers and linguists to most accurately describe the dialect via minimal pairs~\citep{demszky-etal-2021-learning}. Additionally, it opens the potential for TADA to be used for non-English dialects, related languages, and codeswitched variants where small amounts of manually translated data already exists~\citep{diab2010colaba, salloum-habash-2013-dialectal, klubicka-etal-2016-collaborative, costa-jussa-2017-catalan, costa-jussa-etal-2018-neural, popovic2020neural, chen2022calcs, agarwal2022cst5, hamed-etal-2022-arzen}  Furthermore, using a small amount of data, in combination with a parameter-efficient method, reduces compute costs as a barrier for dialect speakers to develop and own language technology within their communities~\citep{ahia-etal-2021-low-resource}.

\subsection{Contrastive Sequence Alignment}
Multilingual NLP has shown that $L_2$ alignment on small amounts of data can provide competitive performance gains to augmentation using translated data during finetuning~\citep{xnli}. This operates on the intuition that similar input representations are likely to lead to similar outputs.

TADA extends this approach to dialects by minimizing the $L_2$ distance between a frozen representation of an SAE input $\vec{CLS}_{sae}$ and the TADA representation of a non-SAE input $\vec{CLS}_{dial}$:
\begin{equation}\label{seq-loss}
    L_{seq} = |\vec{CLS}_{sae} - \vec{CLS}_{dial}|_2
\end{equation}

\subsection{Adversarial Morphosyntactic Alignment}
Since our translated data is aligned at the sequence level, the contrastive loss is only applied to the $\vec{CLS}$ representations. However the variation, and therefore our ideal alignment procedure, operates at the morphosyntactic level.

Lacking token-level aligned data, we instead pursue morphosyntactic alignment using unsupervised adversarial alignment methods~\citep{zhang-etal-2017-earth, lample2018word}. Since our goal is to capture morphosyntactic differences, we use an adversary which pools the entire sequence using a single-layer transformer~\citep{vaswani2017attention} with a two-layer MLP scoring head. A transformer adversary has the expressive capacity to identify misalignment in both individual tokens and their relationships. 

We leave the source dialect frozen which has been shown in computer vision to lead to representations that are composable with downstream modules~\citep{judy-align}. Given the adversarial scoring network $\adv$, a frozen SAE representation $\vec{SAE}$, and a Non-SAE representation after TADA $\vec{Dial}$, we train $\adv$ to maximize:
\begin{equation}
   L_{adv} = \adv(\vec{Dial}) - \adv(\vec{SAE})
\end{equation}

Then, define the morphosyntactic loss for TADA by minimizing the critic loss from $\adv$:
\begin{equation} \label{adv-loss}
   L_{ms} = -\adv(\vec{Dial}) 
\end{equation}

\subsection{Plug-And-Play Application}

Finally, we propose a procedure for applying TADA to downstream tasks. We use composable invertible adapters~\citep{mad-x} as our starting point. Using the 1,000 sentences from WiC, we train these adapters to minimize the combined contrastive and adversarial loss functions: 
\begin{equation}
   L_{TADA} = L_{seq} + L_{ms}
\end{equation}

At test time TADA modules can be stacked behind traditional task adapters~\citep{houlsby2019parameter}. TADA serves to directly align the representations of Non-SAE inputs to the SAE embedding space that these task adapters were trained on. Our experiments show that this consistently improves adapter performance without further training.

\section{Evaluating TADA}
We benchmark TADA on 4 VALUE~\citep{value, multivalue} transformed versions of the GLUE Benchmark~\citep{Wang2018GLUEAM}. As discussed in our limitations, these benchmarks are artificial but enable the evaluation of TADA across multiple tasks and dialects. First, we show how TADA compares to SAE models and task-specific baselines for African American English (AAE). Then, we show that TADA is effective across 4 global dialects of English. Finally, we perform an ablation to evaluate the contribution of each loss function.

For all TADA experiments, we train using 1,000 WiC sentences as described in Section \ref{synth}. We train for 30 epochs with early stopping based on the lowest contrastive loss on a development set of 100 held-out WiC sentences. In Section \ref{hparams}, we report full hyperparameters along with the training details for SAE and VALUE models.

\section{Training Details}\label{hparams}

TADA is trained with the ADAM optimizer for 30 epochs with batch size of 16 and with a hyperparameter search of 5e-4. We keep the model and epoch with lowest $L_2$ loss on the 100 held-out examples. Training takes approx. 30 minutes on an Nvidia GeForce RTX 2080 Ti.

To find this hyperparameter setup, we performed a grid search over batch sizes from ${8, 16, 32}$ and learning rates from ${5\cdot10^{-3}, 5\cdot10^{-4}, 5\cdot10^{-5}}$ for AAVE and used the configuration with the lowest $L_2$ loss on the 100 held-out examples.

For all SAE and VALUE GLUE models, we finetune RoBERTa base for 10 epochs with the ADAM optimizer, a learning rate of $2\cdot10^{-5}$, a batch size of 16, and a linear learning rate warm-up of 6\%. For all SAE and VALUE GLUE adapters, we finetune the original adapter architecture~\citep{houlsby2019parameter} inside RoBERTa base for 20 epochs with the ADAM optimizer, a learning rate of $1\cdot10^{-4}$, a batch size of 16, and a linear learning rate warm-up of 6\%. Training all baseline models took approx. 3 days on an Nvidia GeForce RTX 2080 Ti. Additionally, we report experimental results on the BERT-base model in Appendix \ref{bert-table}.

\subsection{TADA vs. Task-Specific}
\begin{table*}
\centering
\setlength{\tabcolsep}{3pt}
\resizebox{0.99\textwidth}{!}{%
\begin{tabular}{c|cc|cc|cc|cc|cc|cc|cc|cc}
             & \multicolumn{2}{c|}{CoLA} & \multicolumn{2}{c|}{MNLI} & \multicolumn{2}{c|}{QNLI} & \multicolumn{2}{c|}{RTE} & \multicolumn{2}{c|}{QQP} & \multicolumn{2}{c|}{SST2} & \multicolumn{2}{c|}{STSB} & \multicolumn{2}{c}{Mean} \\ 
Test Dialect & Orig.      & TADA     & Orig.      & TADA     & Orig.      & TADA     & Orig.     & TADA     & Orig.     & TADA     & Orig.      & TADA     & Orig.      & TADA     & Orig.      & TADA     \\ \hline
SAE          & 58.3          & 87.2     & 87.2          & 87.2     & 93.2          & 93.2     & 70.8         & 70.8     & 93.9         & 93.9     & 90.5          & 90.5     & 90.5          & 90.5     & 83.5          & 83.5     \\ \hdashline
AAVE         & 14.1          & 29.5     & 83.7          & 84.8     & 90.3          & 91.7     & 67.1         & 67.1     & 86.8         & 88.1     & 92.1          & 91.9     & 88.7          & 89.6     & 74.7          & 77.5 \up{2.8}     \\
Indian       & 16.4          & 15.0     & 82.6          & 83.6     & 89.1          & 90.3     & 66.8         & 66.8     & 86.4         & 87.0     & 90.9          & 91.1     & 88.5          & 88.9     & 74.4          & 74.7 \up{0.3}    \\
Nigerian     & 23.7          & 27.2     & 84.3          & 84.8     & 91.2          & 91.1     & 65.0         & 64.6     & 88.2         & 88.2     & 92.2          & 92.1     & 89.3          & 88.7     & 76.3          & 76.7 \up{0.4}     \\
Singaporean  & -0.4          & 20.3     & 81.4          & 83.0     & 87.7          & 89.3     & 63.2         & 64.3     & 85.2         & 87.3     & 90.9          & 91.1     & 88.1          & 88.5     & 70.9          & 74.8 \up{3.9}    
\end{tabular}
}
\caption{\textbf{Multi-Dialectal} evaluation results across all Tasks (Matthew's Corr. for CoLA; Pearson-Spearman Corr. for STS-B; Accuracy for all others) for 4 Non-SAE Dialect Variants of GLUE created using Multi-VALUE.}
\label{tab:performance}
\end{table*}

Since ours is the first work to attempt task-agnostic dialect adaptation, we benchmark TADA in comparison to prior task-specific methods in Table \ref{main-table}. 

We first establish pure SAE baselines for both full finetuning and adapter training~\citep{houlsby2019parameter}. Interestingly, the gap between SAE performance and AAE performance is similar for adapters (-8.8) and full finetuning (-8.9) when trained on SAE. The minimal effects of the limited capacity of adapters on disparity indicate that dialectal discrepancy is largely within the pretrained LLM before finetuning. Without mitigation, SAE models alone perform poorly on non-SAE input.

We then train two task-specific dialect mitigation following the approach of VALUE, which augments training data with pseudo-dialect examples during finetuning. This is a strong baseline, as it allows the model to adapt specifically to in-domain augmented examples rather than the general sentences used to align TADA modules. When trained on augmented data, adapters (80.7 Avg.)\footnote{Avg. refers to the mean performance across GLUE tasks.} seem to outperform full finetuning (77.5 Avg.). We hypothesize that random initialization of adapters prevents conflicting gradients across dialects which can lead to negative transfer~\citep{wang-etal-2020-negative}.

Finally, we combine TADA with task-specific SAE modules for our task-agnostic approach. TADA succeeds in our goal of generalizable performance improvements, yielding improved robustness for 6 out of 7 tasks for an average increase of 2.8 points on the GLUE benchmark. However, TADA performs 4\% worse on average than task-specific VALUE-augmented adapters. These adapters are trained on larger amounts of dialectal training data directly from each task than TADA, which likely explains their superiority. However, as noted in the table these approaches scale training and storage linearly with the number of tasks, while TADA requires only a constant overhead.

These results are the first to indicate the possibility of task-agnostic dialect adaptation. While performance lags behind the task-specific intervention, these results indicate similar quality is possible with vastly improved scalability. This scalability across tasks is key to truly addressing dialect disparities as NLP has a growing impact across a larger number of tasks.

\begin{table*}
\centering
\begin{tabular}{l|c|c|c|c|c|c|c|c}
           & \multicolumn{8}{c}{AAE Glue Performance}               \\ 
Method     & COLA & MNLI & QNLI & RTE  & QQP  & SST2 & STS-B & Mean \\ \hline
TADA & 29.5 & 84.8 & 91.7 & 67.1 & 88.1 & 91.9 & 89.6  & 77.5 \\ \hdashline
\hspace{5pt}$-L_{ms}$ (Eq. \ref{adv-loss})    & 29.1 & 85.0 & 91.5 & 66.1 & 88.0 & 91.6 & 89.4  & 77.2 \drop{0.3}\\
\hspace{5pt}$-L_{seq}$ (Eq. \ref{seq-loss})   & 0.0  & 31.8 & 50.5 & 36.8 & 47.3 & 50.9 & 10.7  & 32.6 \drop{44.9}
\end{tabular}
\caption{\textbf{TADA Loss Ablation} results for RoBERTa Base for the 7 GLUE Tasks (Matthew's Corr. for CoLA; Pearson-Spearman Corr. for STS-B; Accuracy for all others) for African-American English. Our results show that the combined loss functions of TADA lead to the strongest results.}
\label{tab:ablation}
\end{table*}

\subsection{Cross-Dialectal Evaluation}
We then confirm that TADA generalizes across regional dialects using 3 global dialect translations introduced from \citet{multivalue} in Table \ref{tab:performance}. Beyond AAE, we select Nigerian English and Indian English as they are each estimated to have over 100 million English speakers\footnote{Speaker estimates from the Oxford English Dictionary \href{https://public.oed.com/blog/introduction-to-nigerian-english/}{Introduction to Nigerian English} and the \href{https://censusindia.gov.in/census.website/data/LSI}{Indian Census}.},  Singaporean English as it was identified as particularly challenging.

Despite not explicitly encoding any linguistic features, TADA is not dialect-agnostic. TADA improves average performance by +2.8, +0.3, +0.4, and +3.9 respectively for African American, Indian, Nigerian, and Singaporean Englishes.  

Ultimately, this applicability across dialects reinforces TADAs potential as a general tool, but with key limitations at fully removing the dialect gap. Truly dialect-robust NLP requires generalization across both tasks and dialects, making measuring the performance of both essential. We recommend future works on dialect modeling evaluate both.

\subsection{Ablation Study}
Finally, we show the resuilts from an ablation in Table \ref{tab:ablation} to evaluate the contributions of each loss function to the final TADA methods.
Contrastive loss alone yields close performance to TADA; it consistently underperforms the combined loss functions on 6 out of 7 tasks (-0.3 Avg.). This extends evidence for the efficacy of this simple loss function from the multilingual~\citep{xnli} to the dialectal domain. 

When contrastive loss is removed, the adversarial loss quickly becomes unstable and suffers from mode collapse. This leads to pathological results, with the resulting adapters harming performance for all tasks (-44.9 Avg.).

\section{Conclusions}
English dialects are underserved by NLP, but are both tractable targets for transfer learning and have huge speaking populations~\citep{local-languages}. Models which serve English speakers inherently serve a global population who use the language natively and as a second tongue.

However, current approaches to improve dialectal robustness in English have so far focused only on one task at a time. The scalability of these task-specific methods limits their impact as language technology applications become increasingly diverse and pervasive. We argue that task-agnostic dialectal methods are a clear, yet unexplored path to serve these communities effectively.

We propose a simple yet effective technique TADA to address this, utilizing morphosyntactic data augmentation and alignment loss at both the sequence and morphosyntactic level to train adapter modules. When composed with SAE task adapters, TADA modules improve dialectal robustness consistently on the multi-task GLUE benchmark. Future work should work to further reduce the dialect discrepancy to create more inclusive and equitable English language technology.

\section*{Limitations}
TADA makes use of the pseudo-dialectal translation systems of prior work~\citet{value, multivalue}. We rely on them as they are validated by dialect speakers and have been shown to be predictive of performance on Gold Dialect data. However, they were designed as stress tests of robustness which isolates morphology and syntax. We are therefore unsure how TADA performs when it faces the topical and register shifts which often are associated with naturally occurring dialects. These limitations are similar to localization issues in translated benchmarks~\citep{moradshahi-etal-2020-localizing}.

In this work, we evaluate TADA on only Encoder-only LLMs. Increasingly, both Encoder-Decoder and Decoder-only models are seeing wide-scale use due to their flexibility~\citep{architecture}. Evaluating TADA and developing alternate tailored task-agnostic methodologies on these alternate LLM architectures is left to future work.

\section*{Ethics Statement}
This work refers to linguist-drawn boundaries around dialects. However, dialects are not monolithic and are used in varied ways across sub-communities of speakers. Readers should therefore not understand TADA to remove discrepancies across all speakers as improvements may vary within subcommunities within a dialect~\citep{koenecke2020racial}.
Additionally, as TADA is task-agnostic, it is especially vulnerable to dual use. To mitigate this, we will release TADA under a license that forbids usage with intent to deceive, discriminate, harass or surveil dialect-speaking communities in a targeted fashion.

\section*{Acknowledgements}  We are thankful to Yanzhe Zhang and the anonymous ACL reviewers for their helpful feedback. 
\bibliography{custom}
\bibliographystyle{acl_natbib}

\appendix

\setcounter{table}{0}
\renewcommand{\thetable}{A\arabic{table}}

\begin{table*}
\resizebox{0.99\textwidth}{!}{%
\begin{tabular}{cccc|c|c|c|c|c|c|c|c}
             \multicolumn{4}{c|}{Dialect Adaptation Details} &           \multicolumn{8}{c}{AAE Glue Performance}               \\ \hline
Approach & Method    & Task-Agnostic  & Dialect Params. & COLA & MNLI & QNLI & RTE  & QQP  & SST2 & STS-B & Mean      \\
\hline
N/A                & Finetuning & \checkmark & 0 &	36.0 & 79.6 & 89.2 & 65.3 & 86.2 & 89.7 & 87.4 & 76.2                 \\
N/A                & Adapters   & \checkmark & 0 & 31.4	&	80.8	&	89.2	&	62.1	&	86.0	&	89.8	&	86.9	&	75.1                 \\ \hdashline
VALUE       & Finetuning & \xmark &$T\times 110M$                 &	36.2	& 83.0	& 89.7	& 61.4	& 88.6	&	89.6	&	88.2	&	76.7 \\
VALUE       & Adapters   & \xmark & $T\times 895K$                    & 	36.3	&	82.0	&	89.5	&	66.8	&	85.6	&	88.8	&	88.5	& 	76.8 \\
\hdashline
TADA               & Adapters  & \checkmark & $895K$                  &	38.3	&	81.5	&	89.0	&	62.1	&	87.0	&	90.0	&	88.0	&	76.6                
\end{tabular}
}
\caption{\textbf{Dialect Adaptation GLUE} results of BERT Base~\citep{bert} for the 7 GLUE Tasks (Matthew's Corr. for CoLA; Pearson-Spearman Corr. for STS-B; Accuracy for all others). $T$  is the number of target tasks for dialect adaptation.}
\label{bert-table}
\end{table*}

\end{document}